\documentclass[conference,compsoc]{IEEEtran}
\IEEEoverridecommandlockouts

\usepackage{dsfont}
\usepackage{graphicx}
\usepackage{algorithm}
\usepackage{algpseudocode}
\usepackage{xspace}
\usepackage{xcolor}
\usepackage{enumitem}
\usepackage{comment}
\usepackage{amsmath}
\usepackage{xurl}
\usepackage{wrapfig}

\usepackage{color}
\usepackage{subcaption}
\usepackage{booktabs}

\usepackage{tcolorbox}

\usepackage{amsfonts}
\usepackage{amssymb}

\newtheorem{assumption}{Assumption}

\newcommand{\noModels}{$19$\xspace}
\newcommand{\noMoE}{$16$\xspace}
\newcommand{\noClosed}{$21$\xspace}

\newcommand{\noPairs}{$342$\xspace}
\newcommand{\noPrompts}{$100$\xspace}
\newcommand{\promptLengths}{$\{4,8,10,12,16,24\}$\xspace}

\newcommand{\noTextsTest}{$37$\xspace}
\newcommand{\noTextsBaseline}{$4$\xspace}
\newcommand{\secpar}[1]{\vspace{5pt}\noindent\textbf{#1.}}

\begin{document}

\date{}

\title{Inferring the Size of Large Language Models From Popular Text Memorization\textsuperscript{*}\thanks{\textsuperscript{*}This work is supported by funds from an NUS Presidential Young Professorship of Prashant Nalini Vasudevan..}}
    
\author{
Ivica Nikoli\'c\\
National University of Singapore\\
\texttt{cube444@gmail.com}}

\maketitle

\begin{abstract}
The parameter counts of the most widely used large language models (LLMs) are often withheld by their developers, leaving model size---a primary reference
point for interpreting capabilities and costs---largely undisclosed. We propose a
black-box method to infer conservative lower bounds on LLM size from generated
text outputs alone, requiring nothing beyond the ability to submit text fragments
and observe next-token predictions. Our approach is grounded in a key observation:
popular, widely-circulated texts---such as classical literature, religious texts,
and foundational documents---are present in virtually every large-scale
pretraining corpus, and how accurately a model predicts the next word across text
fragments of varying length is a reliable signal of how much it has memorized them,
which in turn is fundamentally limited by its total parameter count. We aggregate
this memorization signal across a diverse corpus of texts and fragment lengths into
a single accuracy profile vector per model, and build two complementary inference
methods on top of it: a pairwise statistical test that determines which of two
models is larger, and a scaling-law estimator that extracts a one-dimensional
latent index from these vectors via Principal Component Analysis (PCA) to map the
aggregated signal to a parameter count. Validated on a broad set of open-weight
models, both methods produce accurate and reliable lower bounds. 
When applied to popular closed-weight models, our framework recovers internal 
product hierarchies and reveals a clear divergence in industry scaling strategies: 
while some developers yield significantly higher bounds indicative of large 
generational parameter growth, others operate under strict parameter ceilings, 
demonstrating that hidden design choices can be systematically probed even under 
strict API limitations.
\end{abstract}

\section{Introduction}

Large language models (LLMs) have become deeply integrated into many applications and services. The development of the most advanced systems is heavily driven by major tech companies such as OpenAI, Google, Meta, and Anthropic. The most widely used are general-purpose models, which are trained on vast amounts of text to adapt flexibly to a wide variety of open-ended user instructions and tasks. Despite their widespread use, the most capable models remain closed-weight. Providers share very little about how they are built. While keeping specific architectural choices private is an understandable way to protect commercial interests, it is notable that even the most basic descriptive metric—the total parameter count—is almost always kept secret.

This omission has a large impact because model size is a primary variable for understanding, comparing, and deploying language models. If parameter count cannot be used as a control variable, comparing closed-weight and open-weight models becomes ambiguous. Without knowing the size of a model, it is impossible to determine whether a performance advantage stems from algorithmic breakthroughs, higher quality training data, or simply a massive increase in parameter count. Furthermore, size dictates economics. Parameter count tightly bounds the compute and memory required to serve a model, directly driving API pricing and latency. 
Providers may therefore have strong commercial incentives  to hide this number, as it prevents independent verification  of the computational cost and resource requirements associated with running their systems.

Attempting to infer model size directly from a model's text outputs raises 
substantial methodological challenges. Model providers typically restrict 
interaction to APIs where a user can submit a prompt and inspect the returned 
text, but cannot obtain token probabilities, logits, hidden states, architectural 
metadata, or gradients. This constrained interaction is a limited form of model 
property inference---narrower than full model extraction, where an adversary 
reconstructs weights or a functional replica, but relying on the same premise: 
that a private attribute can be extracted through strategic black-box querying. 
Estimating model size purely from returned text requires isolating a signal 
that correlates strongly with parameter count. However, returned text is 
sensitive to numerous unknown variables, including training data, formatting 
constraints, system prompts, the extent of instruction tuning, and the 
hyperparameters used during deployment. Separating the intrinsic effect of 
model size from these unobserved variables is non-trivial.

For this reason, inferring the size of a closed-weight language model from 
black-box interaction is both a practically relevant and well-posed problem 
formulation. It asks what can be learned about a black-box model from its 
returned text alone, while acknowledging that any such inference must contend 
with multiple unobserved variables and limited access to the system.

\secpar{Contributions}
We make the following contributions:

\begin{enumerate}

\item \textbf{Memorization-based size signal.} We show that aggregating
a model's accuracy at predicting the next word of incomplete sentences
drawn from popular, widely-circulated texts, across many such texts and
sentence lengths yields a high-dimensional vector---an \emph{accuracy
profile}---that contains a strong signal for model size.

\item \textbf{Dual lower-bound inference framework.} We derive two
complementary methods for translating accuracy profiles into conservative
lower bounds on model size: (i)~a \emph{scaling-law estimator},
a PCA projection onto a one-dimensional latent index combined with an
exponential scaling law calibrated on open-weight reference models; and
(ii)~a \emph{pairwise test}, a statistical hypothesis test that
establishes which of two models is larger with a controlled false positive
rate. The final bound for any target model is the maximum of the two.

\item \textbf{Validation across open- and closed-weight models.} We calibrate and validate our methods
on 19 dense open-weight models, where the scaling-law estimator achieves
$R^2 = 0.95$ and the pairwise test achieves around 90\% precision and
recall. We then test the calibrated framework on \noMoE open-weight MoE
models---producing empirically valid lower bounds in all evaluated cases---and on \noClosed
popular closed-weight models from Anthropic, Google, OpenAI, and Alibaba,
obtaining lower bound estimates that vary depending on the 
provider.

\end{enumerate}
\section{Problem}

Model size of a large language model $f$ is measured by its parameter count and denoted with $|f|$. 
It is one of the primary factors shaping model capabilities and a key reference point
for interpreting performance comparisons. For closed-weight models, however,
this quantity is often not disclosed.

The most direct form of the problem is to estimate the size of an
unknown model from its behavior alone.

\begin{tcolorbox}[
    colback=gray!7, colframe=black, boxrule=0.6pt, arc=3pt,
    left=8pt, right=8pt, top=5pt, bottom=5pt,
]
\textbf{Problem 1 (Absolute Size Inference).}
Let $f$ be a language model with unknown parameter count $|f|$. Given
black-box query access to $f$, estimate $|f|$.
\end{tcolorbox}

This is the strongest of the problems we study, and obtaining the model size is  always the ultimate goal. However, size inference may not be possible,
or may not be precise. We therefore also study a  simpler problem that asks
only whether one model is larger than another.

\begin{tcolorbox}[
    colback=gray!7, colframe=black, boxrule=0.6pt, arc=3pt,
    left=8pt, right=8pt, top=5pt, bottom=5pt,
]
\textbf{Problem 2 (Relative Size Inference).}
Let $f$ and $g$ be two language models with unknown parameter counts $|f|$ and
$|g|$. Given a threshold $\tau > 0$, determine whether $|f| > (1+\tau)\,|g|$.
\end{tcolorbox}

Relative size inference is weaker than absolute estimation but remains useful
in practice. If two models perform similarly on a benchmark but one turns out
to have more parameters, the smaller model is arguably preferable---it achieves
comparable results at lower computational cost. Drawing this conclusion requires
knowing which model is larger, even without knowing exact counts.

The problem is formulated with a multiplicative threshold $\tau$ rather than
asking simply whether $|f| > |g|$, because the differences that matter in
practice are not marginal ones. Distinguishing a $30$B model from a $32$B model
is of limited consequence---models so close in parameter count carry essentially
the same computational cost and have broadly comparable capacity. What is
practically relevant are substantially larger gaps: $40\%$, $60\%$ and
above, where the size relationship  affects how performance comparisons
should be interpreted. The threshold $\tau$ makes this target range explicit.
Consequently, if a method cannot reliably resolve very small values of $\tau$,
this is not a fundamental shortcoming: the differences it cannot resolve are
precisely those that are rarely consequential in practice.

We study these problems  under the most restricted, black-box access. A tester may
submit prompts to a model and observe its returned text outputs, but nothing more is
assumed. In particular, the tester does not observe model weights, logits, token
probabilities, hidden states, gradients, architectural metadata, training data,
hyperparameters, or post-training procedures. Nor do we assume the models belong
to the same family or were developed under comparable conditions. 
We assume a non-adaptive adversary that is not intentionally attempting to hide or distort the signals used to infer model size.

\section{Approach}
\label{sec:approach}

\begin{figure}[h]
    \centering
    \includegraphics[width=\linewidth]{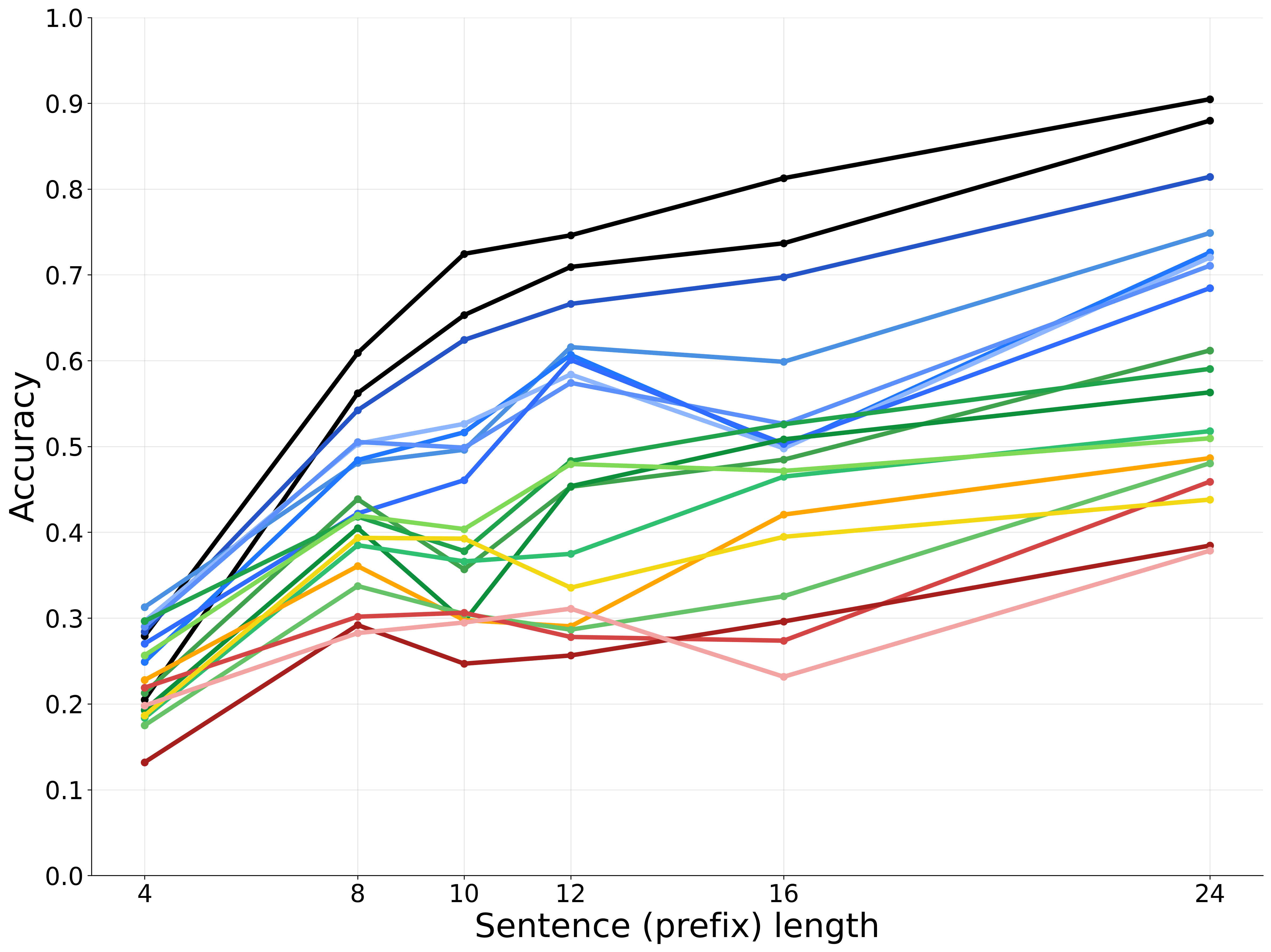}
    \caption{Accuracies of different models on \textit{Alice in Wonderland}.}
    \label{fig:example}
\end{figure}

Our approach is grounded in the key observation illustrated in Figure~\ref{fig:example}, which shows how accurately different large language models predict the next word of incomplete sentences from the book \emph{Alice in Wonderland}, as the length of those sentences varies. 
Colors follow a rainbow gradient, with darker shades indicating larger size models. 
A clear pattern emerges: darker lines tend to appear above lighter ones, revealing that larger models are more accurate at predicting the next word. While individual measurements can be noisy, aggregating next-word accuracy across many sentence lengths and a diverse collection of popular texts yields a stable, robust signal—one that, as we show, correlates strongly with model size.

The following sections develop this methodology in detail for \emph{dense}  models; the implications for Mixture-of-Experts architectures and the conditions under which the approach remains applicable are discussed in Section~\ref{sec:scope}.

\subsection{Notation}
\label{sec:notation}

A \emph{source text} is a canonical, widely-circulated document---a popular
book, religious text, or well-known publication---chosen because such texts
appear in any large-scale pretraining corpus, often multiple times.
A \emph{baseline text} is a relatively obscure work
unlikely to have been seen often enough during pretraining to produce
meaningful memorization effects.

We represent any text as an ordered token sequence
$T = (t_1, t_2, \ldots, t_n)$.
For a language model $M$, a text $T$, and a prefix length $l \geq 1$, the
\emph{next-token accuracy} is
\begin{equation}
  \alpha(M,\,T,\,l)
  \;=\;
  \frac{1}{n - l}
  \sum_{i=l+1}^{n}
  \mathbf{1}\!\left[M(t_{i-l},\ldots,t_{i-1}) = t_i\right],
  \label{eq:alpha}
\end{equation}
where $M(t_{i-l},\ldots,t_{i-1})$ denotes the top-1 token predicted by $M$
given the preceding $l$ tokens. Let $\mathcal{B} = \{B_1,\ldots,B_R\}$ be
the set of baseline texts. The \emph{baseline accuracy} of $M$ at prefix
length $l$ is
\begin{equation}
  \beta(M,\,l)
  \;=\;
  \frac{1}{|\mathcal{B}|}
  \sum_{B\,\in\,\mathcal{B}}
  \alpha(M,\,B,\,l).
  \label{eq:beta}
\end{equation}
We refer to $\alpha(M,T,l)$ as \emph{raw accuracy} and to
$\alpha(M,T,l) - \beta(M,l)$ as \emph{lifted accuracy}. Throughout, we write
$\alpha$ without superscript when the distinction is not relevant, reserving
$\alpha^{\mathrm{r}}$ and $\alpha^{\mathrm{l}}$ for contexts where both are
referenced explicitly. 
We use the hat notation, e.g.\ $\hat{\alpha}$, to
denote subsampled estimators of the corresponding exact quantities.

\subsection{Accuracy as a Signal for Model Size}

The core intuition of our approach is to use a model's next-token accuracy on well-memorized, popular texts to infer its size. In a general setting, next-token accuracy $\alpha(M, T, l)$ depends on many parameters, most of which are unobservable in a black-box environment. We model this dependence as an unknown function $F$:
$$ \alpha(M, T, l) = F(l, T, D_M, \theta_M, \lambda_{M,T,l}) $$
where $\theta_M$ denotes the parameter count of model $M$; $D_M$ represents the training-data distribution (which captures the extent to which text $T$ was seen during pretraining); and $\lambda_{M,T,l}$ captures all other latent factors, including pretraining duration, architecture choices, optimization details, instruction tuning, and deployment-level constraints. 

Because the exact functional form of $F$ is unknown, a difference in raw accuracy between two models could stem from any of its arguments, not just size. To isolate the size parameter $\theta_M$, we must systematically control for the other arguments of $F$. 

We address the dependence on text exposure ($D_M$) through our choice of text, formalized as follows:
\begin{assumption}
For a sufficiently popular and widely-circulated source text $T$, the distribution of $T$ within $D_M$ is approximately the same across models.
\end{assumption}
Virtually all modern large language models are trained on massive web crawls drawn from a common public internet. Therefore, a sufficiently popular document (such as \textit{Alice in Wonderland}) will be heavily and essentially uniformly represented in any large-scale pretraining corpus. 
In Appendix~\ref{sec:appendix:assumption-1}, we provide empirical support for this assump-
tion.
By restricting our queries to these popular texts, we effectively fix the training distribution variable, meaning $D_f \approx D_g \approx D$ for any two models $f$ and $g$. The accuracy dependence on popular texts then simplifies to:
$$ \alpha(M, T, l) \approx F(l, T, D, \theta_M, \lambda_{M,T,l}) $$

With the text exposure essentially held constant across models, the remaining differences in accuracy must stem from only two sources: their size ($\theta_M$) and their latent architectural and training parameters ($\lambda_{M,T,l}$). 

\begin{assumption}Model size is expected to dominate the other latent factors in determining next-token prediction accuracy on highly popular texts.
\end{assumption}
While variations in post-training recipes or specific architectural tweaks ($\lambda$) may affect accuracy, a model's  parameter count ($\theta$) fundamentally determines its intrinsic capacity to accurately predict words from these source texts. Therefore, we posit that the primary determinant of accuracy on well-memorized text is the model size itself. The size factor will overwhelmingly dominate the latent parameters, meaning differences in accuracy across models are driven primarily by differences in size. This insight—that size has a robust, dominant impact on popular text prediction—is the basis of our methodology.

To translate this empirical  signal from accuracy into a reliable measurement of size, we must address noise. A query on a single text may still be skewed by a localized violation of Assumption~1 or by a particular latent factor. Furthermore, computing the exact value of $\alpha(M,T,l)$ requires querying every eligible position in the text, which is computationally expensive. In practice, we therefore estimate this quantity by subsampling a random set of positions $S_{T,l} \subseteq \{l+1,\dots,n\}$:
$$ \hat{\alpha}(M, T, l) = \frac{1}{|S_{T,l}|} \sum_{i \in S_{T,l}} \mathbb{1}[M(t_{i-l}, \dots, t_{i-1}) = t_i] $$

Because individual measurements of $\hat{\alpha}(M, T, l)$ contain variance from subsampling and residual latent noise, we cannot rely on a single measurement. We must aggregate these signals across a broad collection of texts and prefix lengths so that the dominant size signal systematically outweighs the other latent factors.

Both raw and lifted accuracy may contain information about model size, but it is not clear in advance which provides the stronger or more reliable signal. Lifted accuracy, defined as $\hat{\alpha}^{\ell}(M,T,l)=\hat{\alpha}^{r}(M,T,l)-\hat{\beta}(M,l)$, is attractive because it attempts to remove a model's general prediction ability and isolate how much it improves the accuracy beyond the baseline. However, the baseline $\hat{\beta}(M,l)$ is itself only an estimate, computed from a small set of baseline texts, and may therefore have substantial variance; subtracting it can introduce additional noise and sometimes weaken the signal. Raw accuracy $\hat{\alpha}^{r}(M,T,l)$ avoids this estimation error, but it may be more sensitive to other factors unrelated to memorization, such as differences in overall training quality or general language modeling ability, which can raise accuracy on all texts. Since each representation has a plausible advantage as well as a drawback, we use both rather than commit to either one a priori.

\secpar{Accuracy profiles} 
We carry out aggregation and produce accuracy profile vectors. 
Let $\mathcal{T} = \{T_1, \dots, T_K\}$ be a set of source texts (some potentially satisfying Assumption 1), and let $\mathcal{L} = \{l_1, \dots, l_L\}$ be a set of prefix lengths. For each model $M$, we compute both the raw and lifted accuracies for every combination, yielding $2KL$ scalar measurements. We organize these into a unified accuracy profile:
$$ A_{\mathcal{T}, \mathcal{L}}(M) = \left( \hat{\alpha}_r(M, T_j, l_k), \hat{\alpha}_l(M, T_j, l_k) \right)_{\substack{j=1,\dots,K \\ k=1,\dots,L}} \in \mathbb{R}^{2KL} $$
This high-dimensional vector constitutes the accuracy profile $\mathcal{A}_{\mathcal{T},\mathcal{L}}(M)$ on which we build our inference methods.

\subsection{Absolute Size Inference}
\label{sec:absolute-size}

Building on the premise from the previous section---that the accuracy profile $A_{\mathcal{T},\mathcal{L}}(M)$ aggregates distinct measurements to capture the dominant size factor while reducing latent noise---we hypothesize that these high-dimensional profiles share a single, dominant underlying signal that correlates strongly with parameter count $\theta$.

To isolate this signal, we treat the $2KL$-dimensional accuracy profile of each model as a feature vector. For a collection of $N$ models, we form the feature matrix $X \in \mathbb{R}^{N \times 2KL}$. We then apply Principal Component Analysis (PCA) to $X$, retaining only the first principal component to define a one-dimensional latent size index $z_M \in \mathbb{R}$ for each model $M$. 
This score compresses the full accuracy profile---hundreds of individual measurements across texts and prefix lengths---into a single number. This is not an arbitrary compression: by definition of PCA, $z_M$ retains the most variance, meaning it is the single number that best differentiates the models from one another across all accuracy measurements simultaneously.

The question is then what property of the models this differentiating number actually reflects. This is not answered by PCA itself, but by Assumption~2. If the dominant reason why models differ in their accuracy profiles is their parameter count, then the direction that best separates models in profile space is the size direction, and $z_M$ becomes a proxy for $\theta$. Were Assumption~2 false---say, training duration (which is part of $\lambda$) varied wildly across models while size remained roughly constant---the first principal component would capture training duration instead. The validity of using $z_M$ as a size index therefore rests on Assumption~2: that among all the factors shaping accuracy profiles, $\theta$ is the primary source of between-model variation.

The choice to use a simple linear projection method such as PCA rather than a more complex dimensionality reduction technique is driven by the need for robustness. In our setting, the number of features (hundreds of accuracy measurements) far exceeds the number of observations (a limited number of available open-weight models). Complex non-linear methods, such as autoencoders or kernel PCA, require significantly larger sample sizes to reliably map underlying manifolds; applied here, they would most likely severely overfit. Because next-token accuracy generally increases monotonically with scale, linear PCA is sufficient to capture this shared signal, acting effectively as a robust, data-driven weighted average. 

Crucially, relying on a linear projection does not imply that the relationship between next-token accuracy and model size is linear. Standard scaling laws dictate that behavioral capabilities scale non-linearly with parameter count. We account for this by separating the problem into two distinct steps: scoring the model's behavior, and then predicting its size. The linear PCA is used strictly to collapse the multi-dimensional space into a well-behaved, ordinal index of general capability ($z_M$). We then explicitly capture the non-linear relationship between capacity and parameter count \emph{after} the projection, fitting an exponential curve to map $z_M$ to $\theta$.

\secpar{Fitting the Scaling Law}
Once the models are projected onto the 1D latent index $z$, we construct a size estimator 
by using open-weight models as reference anchors. Let $\mathcal{M}_{\mathrm{ref}}$ be a 
set of models whose true parameter counts $\theta_i$ are publicly known.

The choice of functional form is motivated by neural scaling laws~\cite{kaplan2020scaling, 
hoffmann2022training}, which show that model capability improves roughly logarithmically 
with parameter count: each successive gain in performance requires an exponential increase 
in parameters. Since our latent index $z$ is a linear measure of behavioral capability, 
inverting this relationship implies that parameter count grows exponentially with $z$. 
We therefore model:
\begin{equation}
    \hat{\theta}(z) = A \cdot e^{B \cdot z}
\end{equation}

To fit $A$ and $B$, we take the natural logarithm of both sides, converting the exponential 
relationship into a linear one:
\begin{equation}
    \ln(\theta_i) = B \cdot z_i + c, \quad c = \ln(A)
\end{equation}
We then solve for $B$ and $c$ via ordinary least squares regression on the pairs 
$(z_i, \ln(\theta_i))$ from the reference models, and recover $A = e^c$.

The resulting calibrated curve maps any latent score directly to a parameter count estimate. 
For a closed-weight model, we submit black-box queries, compute its accuracy profile, 
project it onto the shared PCA axis to obtain $z_{\mathrm{closed}}$, and evaluate 
$$\hat{\theta}(z_{\mathrm{closed}}) = A \cdot e^{B \cdot z_{\mathrm{closed}}}$$ to produce 
the absolute size estimate.

\subsection{Relative Size Inference}
\label{sec:relative-size}

While absolute size estimation relies on projecting the entire model set to find a
global scaling axis, we can also use the accuracy profiles to evaluate pairwise
comparisons. For two models $f$ and $g$, the entry-wise difference of their accuracy
profiles,
$$D(f, g) = A_{\mathcal{T},\mathcal{L}}(f) - A_{\mathcal{T},\mathcal{L}}(g)$$
captures the accuracy gap at every $(T_j, l_k)$ combination. Under our core
hypothesis, if $\theta_f > \theta_g$, the dominant capacity factor should manifest as
a consistent positive bias across the entries of $D(f, g)$. However, individual
entries remain subject to noise from latent factors $\lambda$ and potential localized
violations of Assumption~1, so we cannot simply operate on this difference vector
directly; we instead aggregate this evidence into a formal statistical test that
provides explicit probabilistic guarantees on the false positive rate.

The key structural challenge in designing this test is that the entries of $D(f,g)$
are not statistically independent. Measurements drawn from the same source text $T_j$
at different prefix lengths $l_k$ share the exact same underlying document exposure.
If one model has a systematic advantage on that document---due to a $\lambda$-driven
anomaly in its training data distribution---it will manifest simultaneously across
multiple prefix lengths. A naive statistical test that treats all entries as
independent would falsely inflate the effective sample size and produce highly
unreliable $p$-values.

We resolve this by employing a one-sided \emph{blocked} sign-permutation test. This non-parametric approach is designed precisely for data with unknown correlation structures within distinct groups. Rather than permuting individual observations, the test permutes entire blocks. In our setting, each source text $T_j$ forms a single block. While measurements \emph{within} a text may be arbitrarily correlated across prefix lengths, different texts represent distinct, exchangeable units sampled from the broader pretraining landscape.

Before applying the test, we summarize the entries of $D(f,g)$ within each block into a single scalar score $s_j$. We average the accuracy differences for text $T_j$ over the $L$ prefix lengths:
$$ s_j = \frac{1}{L}\sum_{k=1}^{L}\left( \hat{\alpha}(f, T_j, l_k) - \hat{\alpha}(g, T_j, l_k) \right) $$

A subtlety is that the test does not explicitly parameterize the size gap $\tau$
introduced in Problem~2. Consequently, we cannot theoretically certify that a
statistically significant accuracy advantage of $f$ over $g$ implies
$\theta_f \geq (1+\tau)\theta_g$ for a prescribed $\tau$. Instead, we treat this as
an empirical question and evaluate the test across a range of $\tau$ values in
Section~4, to determine whether the test is reliable only for certain size gaps (as shown in Section~\ref{sec:relative-size-test}, the test performs well even when the gap $\tau$ is very small, i.e. 1\%).

The null hypothesis posits that these block scores do not systematically favor $f$ over $g$ across the independent texts, while the alternative asserts that they do. Formally,
$$ H_0: \mathbb{E}[s_j] \leq 0, \qquad H_1: \mathbb{E}[s_j] > 0 $$
Under $H_0$, there is no dominant accuracy advantage for $f$, meaning the sign of each text's block score is driven by symmetric noise and is equally likely to be positive or negative. The sign-permutation test enumerates sign-flipping assignments across the text blocks to construct an exact null distribution. It then evaluates how extreme the observed sum of block scores is relative to that distribution. If the observed sum is sufficiently unlikely under the null, we reject $H_0$ and conclude that $f$ exhibits a statistically significant accuracy advantage over $g$, which under our assumptions is interpreted as evidence that $f$ is larger.

\subsection{Scope and Applicability}
\label{sec:scope}

Our methodology fundamentally relies on the premise that large language models are trained on similar, widely circulated documents (Assumption~1), and that the extent to which a model memorizes these texts is inherently dominated by its total parameter count (Assumption~2). Consequently, our approach succeeds when this memorization signal is preserved and accurately measurable, and it breaks down when the signal is absent, overwritten, masked, or somehow decoupled from the total model size. We detail these boundary conditions below.

\secpar{Training Data and Catastrophic Forgetting}
For the behavioral signal to exist, the target model must have been exposed to the standard internet-scale corpora where our source texts reside. If a model was trained on a strictly private or highly specialized dataset, Assumption~1 fails entirely. Furthermore, even if the model was initially exposed to these texts, aggressive fine-tuning can cause catastrophic forgetting. For example, a model heavily specialized for code generation (e.g., OpenAI Codex) may retain little of its general literary knowledge. In such cases, a low prediction accuracy reflects the specialized fine-tuning (captured in our latent factor $\lambda$) rather than a small parameter count $\theta$, thereby diluting the signal. Thus, our approach is strictly scoped to \emph{general-purpose} models whose broad language capabilities have not been degraded by domain-specific specialization, or any other form of extensive fine-tuning. 

\secpar{Strategic Evasion}
We operate in a \emph{non-adaptive} threat model. We assume the target systems are fixed and their training was not intentionally designed to conceal size-relevant evidence. If a provider were to intentionally suppress next-token accuracy on popular texts---for instance, through targeted unlearning of public-domain literature for copyright reasons---the accuracy profile $\mathcal{A}_{\mathcal{T},\mathcal{L}}(M)$ would be artificially deflated. Because this intentionally weakens the memorization signal, our method would underestimate the model's size. We therefore make no claims of robustness against adversarial unlearning or strategic evasion.

\secpar{Parameter Uniformity and Mixture-of-Experts}
Our mapping from next-token prediction accuracy to model size assumes a dense architecture. In a dense model, every parameter participates uniformly in every forward pass, creating a stable, direct relationship between total parameter count $\theta$ and next-token accuracy. Mixture-of-Experts (MoE) architectures break this uniformity. By routing tokens to specialized subsets of experts, the parameters engaged in any given prediction form a complex, input-dependent fraction of the total parameter count $\theta_{\mathrm{total}}$. Consequently, the dense scaling laws cannot yield precise absolute estimates for MoE targets.

However, a conservative lower-bound interpretation remains  sound. Because an MoE prediction utilizes a strict subset of its total parameters, the behavioral capacity observed can never exceed the capacity of a dense model of size $\theta_{\mathrm{total}}$. Applied to an MoE target, our absolute estimator therefore satisfies:
$$ \hat{\theta}(M_{\mathrm{MoE}}) \;\leq\; \theta_{\mathrm{total}}(M_{\mathrm{MoE}}) $$
Similarly, for relative size inference, if the statistical test concludes that an MoE model $f$ is behaviorally larger than a known dense model $g$, we can safely deduce that $\theta_{\mathrm{total}}(f) \geq |g|$. Due to this asymmetry, we exclude MoE models entirely from our scaling law calibration, treating them strictly as evaluation targets to be lower-bounded.

\secpar{Operational Signal Constraints}
Finally, extracting a clean accuracy signal requires the model to adhere to basic operational standards. To allow automated accuracy measurement, the model must possess sufficient instruction-following capabilities to output exactly the immediate next word without appending conversational filler. Furthermore, its safety filters must not be aggressively over-tuned to the point of blocking classic literary completions entirely. We also explicitly require that extended chain-of-thought or reasoning protocols be disabled. If allowed to generate invisible reasoning tokens before answering, a model could artificially inflate its next-token accuracy through intermediate test-time computation, distorting the intrinsic parameter-driven signal we aim to isolate.

\section{Evaluation}
\label{sect:eval}

We evaluate our approach experimentally to assess its performance across different classes of models, including both open-weight and closed-weight architectures. Additional analysis relevant to the validity of Assumption~1 is provided in Appendix~\ref{sec:appendix:assumption-1}. 

\subsection{Framework}

\secpar{Models}
We analyze and query models available through the OpenRouter API~\cite{openrouter}. As stated previously, we only consider general-purpose instruction-following models for which chain-of-thought reasoning can be disabled, either because it is absent by default or because an explicit option to disable it is exposed. 
Some models are excluded due to rate limits they impose or excessive cost.
We select models with at least 8B parameters (for open-weight models). 
When multiple versions of a model family are available, we include at most two: the latest, and some earlier version. As a result, this yields \noModels open-weight dense models, \noMoE open-weight Mixture-of-Experts models, and \noClosed popular closed-weight models.

\secpar{Texts}
In our preliminary tests, we found that \textit{Hamlet}, \textit{The Bible}, and \textit{The U.S. Constitution}, have been well memorized by one of the tested models. To generate more titles in an unbiased way, we provided these three to an unrelated AI system and asked for a list of more such popular public-domain texts. As a result, we obtained \noTextsTest popular source texts for use in the tests, comprising popular book classics, 
religious texts, and widely circulated documents; the full list is given in 
Table~\ref{tbl:novels} of Appendix~\ref{sec:appendix:texts-and-prompts}. 
We stress  that the sets of texts have not been additionally filtered in any other way. 
Furthermore, we collect \noTextsBaseline baseline texts consisting 
of recently published and relatively obscure works. 
Since the number of popular texts used is already relatively small, we do not attempt to reduce it further algorithmically.

\secpar{Prompts and Queries}
From each text, we sample incomplete prefixes of six different lengths \promptLengths chosen as they capture the range over which 
next-token accuracy exhibits large variance across models: accuracy tends to plateau 
beyond 24 tokens, though the exact point varies across texts --- for some texts it occurs 
earlier, for others later. We fix a single shared set of lengths across all texts to avoid 
any risk of overfitting the evaluation to particular documents. For each length, we 
sample \noPrompts incomplete prefixes at uniformly random positions. This number is chosen to balance two competing concerns: 
enough samples to reduce the variance of the accuracy estimate to a manageable level, while 
keeping the total number of API queries affordable, since each query incurs a cost. Each 
prefix is queried using five different prompt templates to avoid favoring models that are more sensitive to a particular phrasing. These prompts are given in 
Table~\ref{tbl:prompts} of Appendix~\ref{sec:appendix:texts-and-prompts}. A prediction at a given position is counted as correct if any of 
the five queries returns the correct next word. The accuracy for each (model, text, prefix 
length) triple is then computed as the fraction of correct positions out of \noPrompts.
To create an accuracy profile vector for each model (of dimension $2 \cdot 6 \cdot \noTextsTest = 444$),  we need to query the model with  $100 \cdot 6 \cdot 5 \cdot (\noTextsTest +  \noTextsBaseline)=123000$ different prompts, which corresponds to a few million tokens of input text and usually less than a million of output tokens.

\subsection{Open-Weight Dense Models}
\label{sec:eval_dense}

We first validate our methodology exclusively on open-weight dense models. As discussed in Section 3.5, dense models possess a clearly defined total parameter count $\theta$ that directly influences every forward pass, allowing us to establish a baseline. Our evaluation dataset consists of \noModels dense models given in Table~\ref{tbl:models-open-dense}.
\begin{table}[!t]
\caption{Dense open-weight models, sorted by size, expressed in billions of parameters.}
\label{tbl:models-open-dense}
\centering
\normalsize
\setlength{\tabcolsep}{3pt}
\begin{tabular}{p{0.72\columnwidth}r}
\toprule
Model & Size  \\
\midrule
nousresearch/hermes-3-llama-3.1-405b & 405 \\
nousresearch/hermes-4-405b & 405 \\
mistralai/mistral-large-2411 & 124 \\
cohere/command-a & 111 \\
cohere/command-r-plus-08-2024 & 104 \\
qwen/qwen-2.5-72b-instruct & 72 \\
meta-llama/llama-3.1-70b-instruct & 70 \\
meta-llama/llama-3.3-70b-instruct & 70 \\
cohere/command-r-08-2024 & 35 \\
google/gemma-2-27b-it & 27 \\
google/gemma-3-27b-it & 27 \\
qwen/qwen3.5-27b & 27 \\
mistralai/mistral-saba & 24 \\
mistralai/mistral-small-3.2-24b-instruct & 24 \\
microsoft/phi-4 & 14 \\
google/gemma-3-12b-it & 12 \\
mistralai/mistral-nemo & 12 \\
qwen/qwen3.5-9b & 9 \\
meta-llama/llama-3.1-8b-instruct & 8 \\
\bottomrule
\end{tabular}
\end{table}

\subsubsection{Absolute Size Inference}

\begin{figure}[t]
    \centering
    \includegraphics[width=\linewidth]{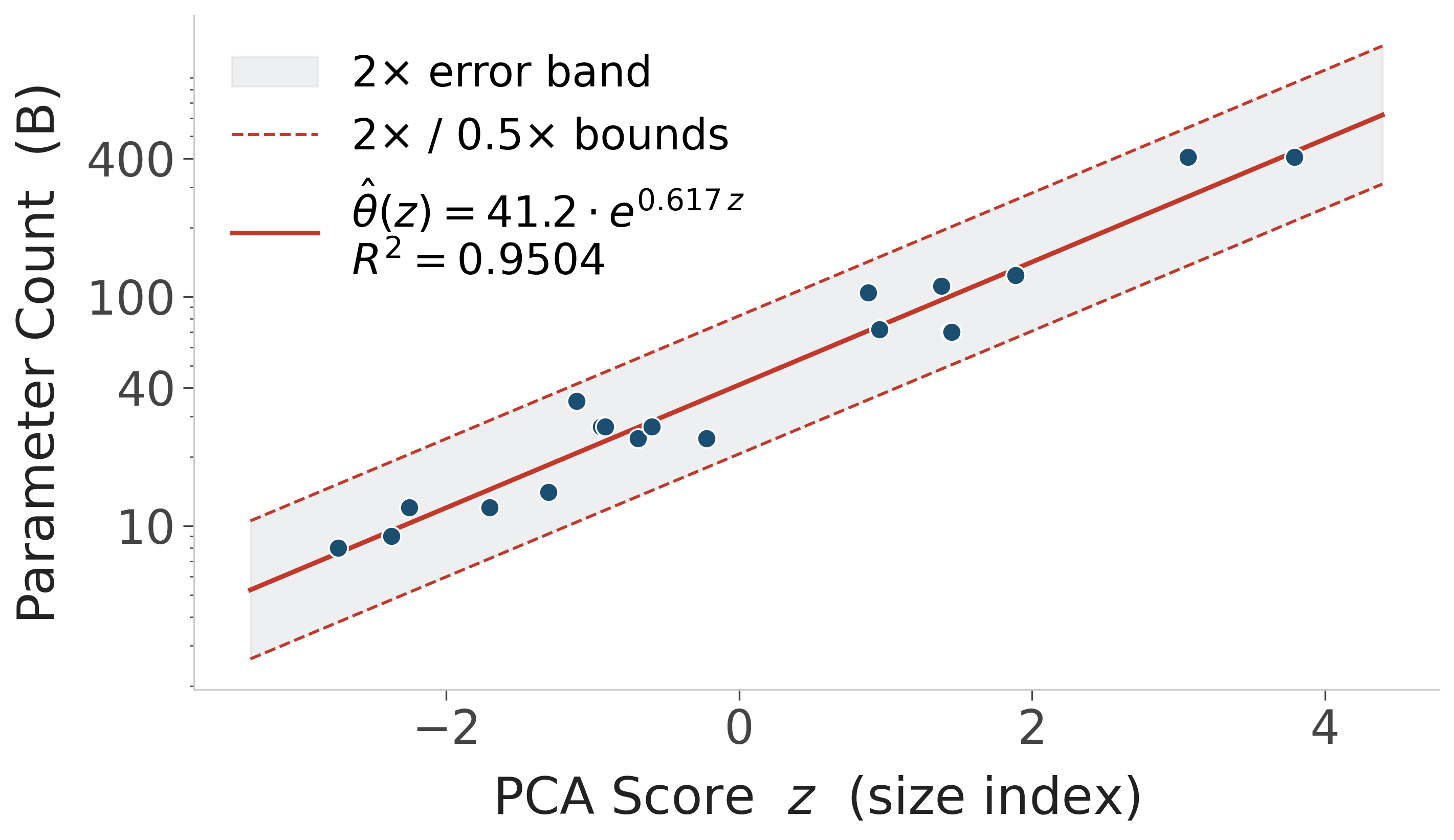}
    \caption{Fitted exponential scaling law $\hat{\theta}(z) = 41.2 \cdot e^{0.617z}$ mapping
the PCA latent size index~$z$ to true parameter count (log scale). Each point
corresponds to one of the 19 open-weight dense models.}
    \label{fig:pca}
\end{figure}

As stated in Section~\ref{sec:absolute-size}, we stack the accuracy profile vectors for all models and construct a matrix and then project it to a 1-dimensional latent size index $z$ using Principal Component Analysis. Then we fit the exponential scaling law $\hat{\theta}(z) = A \cdot e^{B\cdot z}$ to the full set of \noModels models. The ordinary least squares regression yields the parameters:
$$ \hat{\theta}(z) = 41.18 \cdot e^{0.62 \cdot z} $$
This empirical curve achieves a very high goodness-of-fit $R^2 = 0.9504$, refer to Figure~\ref{fig:pca}. This strong correlation provides empirical support for our core hypothesis: despite potentially significant variations in pretraining configurations, optimizers, and instruction-tuning recipes across different developer families, the overarching capacity to recall popular texts remains dominated by parameter count. 

While the global fit confirms the underlying trend, estimating the size of an unseen closed-weight model requires the projection axis and scaling law to generalize to data outside the calibration set. To simulate this, we employ leave-one-out cross-validation (LOO-CV). Under LOO-CV, a target model is completely removed from the dataset; the PCA axis is recomputed, the remaining 18 models are projected to fit a new scaling curve, and the held-out model's size is ultimately inferred.
Cross-validation yields an independent validation score of $R^2 = 0.9387$. Crucially, $100\%$ of the predictions fall within a  factor-of-two error band ($0.5\times \leq \text{Predicted} \leq 2.0\times$ True Size). 

This bounded error margin becomes essential when we later evaluate  Mixture-of-Experts models. Rather than relying on the exact point estimate $\hat{\theta}(z)$, our objective for these models is to establish a minimum size that parallels the lower bounds generated by our relative size test. By incorporating the empirically observed floor and adopting $0.5 \cdot \hat{\theta}(z)$, we use the absolute test point estimate to derive a conservative lower bound.

%
%
\begin{table*}[t]
\caption{Open-weight Mixture-of-Experts models, their actual sizes and our lower-bound (LB) estimates (expressed in billions of parameters). \textit{Best LB} denotes our predicted lower bound, derived as the maximum of our two inference methods: the absolute scaling-law projection (\textit{Abs. LB}) and the relative pairwise permutation tests (\textit{Rel. LB}). \textit{Abs. Size} denotes the actual point estimate evaluated from the empirical curve, bounded down by 0.5$\times$ inside \textit{Abs. LB}. \textit{Tightness} indicates the ratio of the predicted best lower bound to the actual size, expressed as a percentage. \textit{Source} denotes whether the \textit{Best LB} was yielded by the absolute size  (Abs) or relative size (Rel) test.}
\label{tbl:moe-results-eval}
\centering
\normalsize
\begin{tabular}{l|r|r|rrrrr}
\toprule
Model & Size & Best LB & Tightness & Source & Abs. Size & Abs. LB & Rel. LB \\
\midrule
moonshotai/kimi-k2.5 & 1000 & 405 & 40\% & Rel & 667 & 334 & 405 \\
xiaomi/mimo-v2-pro & 1000 & 111 & 11\% & Rel & 198 & 99 & 111 \\
z-ai/glm-5 & 744 & 222 & 30\% & Abs & 444 & 222 & 124 \\
mistralai/mistral-large-2512 & 675 & 143 & 21\% & Abs & 286 & 143 & 124 \\
deepseek/deepseek-chat-v3.1 & 670 & 405 & 60\% & Rel & 434 & 217 & 405 \\
baidu/ernie-4.5-vl-424b-a47b & 424 & 111 & 26\% & Rel & 141 & 71 & 111 \\
meta-llama/llama-4-maverick & 400 & 111 & 28\% & Rel & 118 & 59 & 111 \\
qwen/qwen3.5-397b-a17b & 397 & 111 & 28\% & Rel & 156 & 78 & 111 \\
z-ai/glm-4.5 & 355 & 124 & 35\% & Rel & 234 & 117 & 124 \\
baidu/ernie-4.5-300b-a47b & 300 & 44 & 15\% & Abs & 89 & 44 & 35 \\
qwen/qwen3-235b-a22b-2507 & 235 & 104 & 44\% & Rel & 90 & 45 & 104 \\
microsoft/wizardlm-2-8x22b & 176 & 151 & 86\% & Abs & 302 & 151 & 124 \\
qwen/qwen3.5-122b-a10b & 122 & 35 & 29\% & Rel & 64 & 32 & 35 \\
meta-llama/llama-4-scout & 109 & 35 & 32\% & Rel & 37 & 18 & 35 \\
z-ai/glm-4.5-air & 106 & 35 & 33\% & Rel & 47 & 23 & 35 \\
qwen/qwen3-next-80b-a3b-instruct & 80 & 35 & 44\% & Rel & 51 & 25 & 35 \\
\bottomrule
\end{tabular}
\end{table*}

\subsubsection{Relative Size Inference}
\label{sec:relative-size-test}

We run the relative size test on all \noPairs ordered model pairs constructed
from the \noModels different dense models, 
and in each statistical hypothesis test we use the conventional significance level $\alpha_{sig}=0.05$ (i.e., a 5\% chance of incorrectly concluding that one model is larger when it is not).
As the test produces a binary decision for each pair of models, we
evaluate it using standard precision, recall, and accuracy metrics.
Ground truth is determined by the threshold parameter $\tau$: model $f$ is
considered truly larger than model $g$ if $|f| > (1+\tau)|g|$.
\begin{figure}[t]
    \centering
    \includegraphics[width=\linewidth]{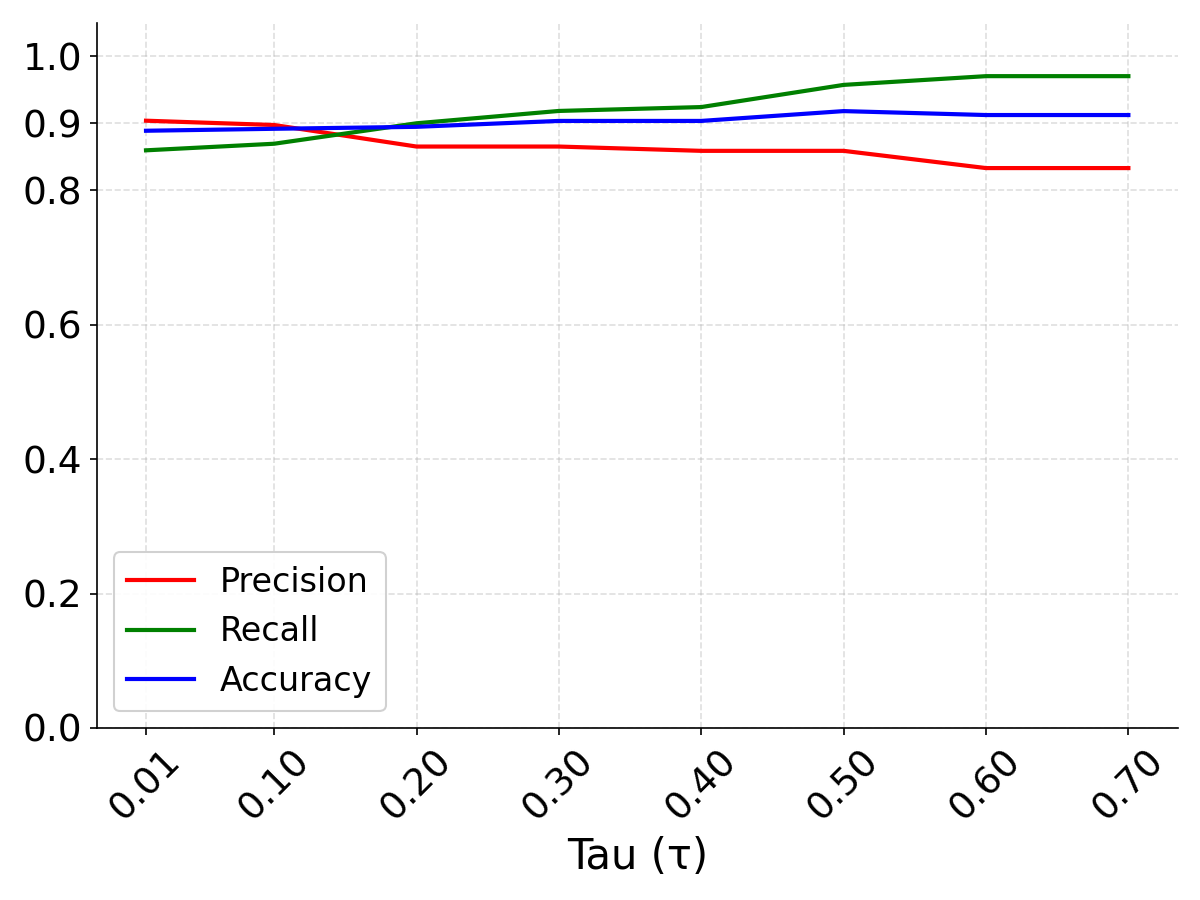}
    \caption{Precision, recall, and accuracy of the relative size test with respect to different values of $\tau$, for $\alpha_{sig}=0.05$.}
    \label{fig:performance}
\end{figure}

Figure~\ref{fig:performance} reports these metrics as functions of the size threshold $\tau$. 
Precision---the fraction of positive decisions that are correct--- starts at $0.91$ and slowly drops, but remains above $0.83$ across the entire range, while recall---the fraction of truly
larger pairs that the test detects---rises from $0.87$ at $\tau=0.01$ to
$0.97$ at $\tau=0.70$, reflecting that larger gaps $\tau$ produce a stronger,
more consistent size signal.
Importantly, even at $\tau=0.01$---with ground truth basically being $|f|>|g|$---the test is already strong, with approximately 90\% of all positive decisions being correct, and a similar percentage for the fraction of truly positive pairs detected by the test.
We therefore conclude that our relative-size test can determine whether  $|f|>|g|$.

%
%
\begin{table*}[t]
\caption{Closed-weight models and our lower-bound (LB) size estimates (expressed in billions of parameters). 
\textit{Best LB} denotes our predicted lowest allowable parameter count, derived as the maximum of the absolute scaling-law projection (\textit{Abs. LB}) and the relative pairwise permutation tests (\textit{Rel. LB}). \textit{Abs. Size} denotes the actual point estimate evaluated from the empirical curve, bounded down by 0.5$\times$ inside \textit{Abs. LB}. \textit{Source} denotes whether the \textit{Best LB} was produced by the absolute size (Abs) or relative size (Rel) test.}
\label{tbl:closed-results-eval}
\centering
\normalsize
\begin{tabular}{l|r|rrrr}
\toprule
Model & Best LB & Source & Abs. Size & Abs. LB & Rel. LB \\
\midrule
anthropic/claude-opus-4.6 & 433 & Abs & 866 & 433 & 405 \\
anthropic/claude-sonnet-4.6 & 405 & Rel & 563 & 281 & 405 \\
anthropic/claude-sonnet-4 & 405 & Rel & 572 & 286 & 405 \\
anthropic/claude-haiku-4.5 & 49 & Abs & 99 & 49 & 35 \\
anthropic/claude-3.5-haiku & 35 & Rel & 51 & 25 & 35 \\
google/gemini-3-flash-preview & 405 & Rel & 638 & 319 & 405 \\
google/gemini-2.5-flash & 104 & Rel & 144 & 72 & 104 \\
google/gemini-3.1-flash-lite-preview & 124 & Rel & 207 & 103 & 124 \\
google/gemini-2.5-flash-lite & 27 & Rel & 36 & 18 & 27 \\
openai/gpt-5.3-chat & 217 & Abs & 435 & 217 & 124 \\
openai/gpt-5.4 & 111 & Rel & 163 & 81 & 111 \\
openai/gpt-4.1 & 111 & Rel & 122 & 61 & 111 \\
openai/gpt-4o & 111 & Rel & 139 & 69 & 111 \\
openai/gpt-5.4-mini & 12 & Rel & 21 & 11 & 12 \\
openai/gpt-4.1-mini & 12 & Rel & 21 & 10 & 12 \\
openai/gpt-3.5-turbo & 9 & Rel & 17 & 9 & 9 \\
openai/gpt-5.4-nano & 2 & Abs & 4 & 2 & --- \\
openai/gpt-4.1-nano & 3 & Abs & 7 & 3 & --- \\
qwen/qwen3.6-max-preview & 104 & Rel & 96 & 48 & 104 \\
qwen/qwen3.6-flash & 35 & Rel & 32 & 16 & 35 \\
qwen/qwen3.5-flash-02-23 & 27 & Rel & 25 & 13 & 27 \\
\bottomrule
\end{tabular}
\end{table*}
\subsection{Open-Weight Mixture-of-Experts Models}
\label{sec:eval_moe}

In this section, we evaluate our approach on 16 open-weight Mixture-of-Experts (MoE) models, which are detailed in Table~\ref{tbl:moe-results-eval}.
As discussed in Section~3.5, MoE architectures break the parameter uniformity assumption because only a subset of parameters is active during any given prediction. Consequently, the dense scaling laws cannot yield precise point estimates for MoE targets. Instead, we use the previously evaluated dense models from Section~\ref{sec:eval_dense} as baselines to establish lower bounds on the total parameter counts of the MoE models.

To achieve this, we leverage both of our proposed methodologies to generate two distinct lower bound estimates for each target MoE model. For the absolute lower bound, we project the MoE model's accuracy profile onto the established latent size index $z$ and evaluate the fitted scaling law $\hat{\theta}(z) = 41.18 \cdot e^{0.62 \cdot z}$ to produce a point estimate (\textit{Abs. Size}). Crucially, to account for empirical variance and architectural differences, we apply the conservative $0.5\times$ multiplier—established by the bounds of our LOO-CV evaluation in Section~\ref{sec:eval_dense}—to this point estimate to define an  absolute test lower bound (\textit{Abs. LB}). For the relative lower bound (\textit{Rel. LB}), we apply the pairwise statistical permutation test between the target MoE model and all dense baseline models, taking the best bound found: the parameter count of the largest dense model over which the MoE model demonstrates a statistically significant advantage. Our final inferred lower bound (\textit{Best LB}) for each target MoE model is taken as the maximum of these two estimates.

The outputs of this approach, presented in Table~\ref{tbl:moe-results-eval}, demonstrate that the inferred \textit{Best LB} is  valid across all \noMoE evaluated MoE models, i.e. for every model the reported \textit{Best LB} does not exceed the model's true parameter count.
Furthermore, the results highlight that both inference methods actively contribute to extracting the most informative bounds. The column \textit{Source} shows that neither method strictly dominates the other; rather, the final \textit{Best LB} is frequently established by the absolute size test for some models and by the relative size test for others. 
The combination of generalized curve projection and head-to-head statistical testing ensures that we capture the highest stable lower bounds possible.

Despite successfully yielding reliable minima, the tightness of these predictions relative to the true model size varies significantly and unpredictably. For some models, the valid predicted bound captures a large fraction of the total size—such as $86$\% for \texttt{microsoft/wizardlm-2-8x22b} or $60$\% for \texttt{deepseek/deepseek-chat-v3.1}. For others, the estimate falls far short of the actual capacity; \texttt{xiaomi/mimo-v2-pro}, for example, is bounded at $111$B despite having $1000$B total parameters (an $11$\% tightness factor). This wide variation illustrates that there is no obvious  multiplier linking these  lower bounds to total parameters for MoE systems. Ultimately, while empirically our framework provides reliable lower bounds, extracting a precise  parameter count strictly from these bounds remains infeasible for MoE without further architectural disclosures.

\subsection{Closed-Weight Models}
\label{sec:eval_closed}

In this section, we apply our framework to \noClosed popular closed-weight models\footnote{Certain prominent models are excluded from our evaluation: Anthropic Claude Opus 4 (high API cost), Google Gemini Pro (inability to disable reasoning),  OpenAI \textit{o}-series (inability to disable reasoning), OpenAI \textit{pro}-series and Turbo 4 (high API cost), Qwen 3.5 Plus (API rate limits).},  detailed in Table \ref{tbl:closed-results-eval}, including systems from Anthropic, Google, OpenAI, and Alibaba. Because these models are closed-weight, their exact architectural details remain private, and we do not know whether they rely on dense or Mixture-of-Experts (MoE) architectures. As a conservative measure, we treat every closed-weight model as a potential MoE and, following the methodology established in Section \ref{sec:eval_moe}, use both inference methods to establish lower bounds on their parameter counts, reporting the higher of the two as our final inferred lower bound.

Table \ref{tbl:closed-results-eval} presents the lower-bound size estimates for all \noClosed closed-weight models. The inferred bounds reveal distinct scaling strategies across developers while consistently reflecting internal product hierarchies. Let us analyze these results separately by provider.

\secpar{Anthropic}
Within the Claude 4 family, the framework assigns its highest overall estimate to \texttt{claude-opus-4.6}, yielding a lower bound of 433B parameters. The two Sonnet variants, \texttt{claude-sonnet-4} and \texttt{claude-sonnet-4.6}, yield closely matching estimates for both relative size bounds (405B) and absolute size bounds (281B and 286B). While the Sonnet bounds converge closely with the flagship Opus, this apparent ceiling likely stems from the absence of a larger open-weight dense baseline in our evaluation set; the availability of such a model would presumably allow the relative-size test to successfully establish a higher lower bound for Opus. Within the Haiku product line, \texttt{claude-3.5-haiku} is bounded at 35B, while \texttt{claude-haiku-4.5} yields an increased lower bound of 49B.

\secpar{Google}
The Google models exhibit pronounced generational scaling within their respective tiers. The \texttt{gemini-2.5-flash} model is bounded at 104B, while its successor \texttt{gemini-3-flash-preview} jumps to 405B. Likewise, \texttt{gemini-2.5-flash-lite} yields a bound of 27B, whereas \texttt{gemini-3.1-flash-lite-preview} reaches 124B. This approximately fourfold increase across both tiers is atypical for incremental version updates. It suggests either a large increase in total parameters between the 2.5 and 3.x generations, or a fundamental architectural shift—such as transitioning toward denser computation or significantly increasing the active routing capacity per token in an MoE framework.

\secpar{OpenAI}
Surprisingly, the largest lower bound among OpenAI models is assigned to \texttt{gpt-5.3-chat}, at 217B—roughly twice the bound of the next closest models. Notably, \texttt{gpt-4o} aligns closely with those next closest models, yielding the same 111B lower bound as \texttt{gpt-5.4} and \texttt{gpt-4.1}. The lightweight variants, \texttt{gpt-5.4-mini} and \texttt{gpt-4.1-mini}, both yield lower bounds of 12B, while the nano-scale variants \texttt{gpt-5.4-nano} and \texttt{gpt-4.1-nano} sit at 2B and 3B respectively. The identical inferred bounds between the 4.1 and 5.4 prefix generations indicate a strict parameter ceiling. This suggests OpenAI has perhaps effectively separated performance improvements from parameter expansion, driving capability gains primarily through better algorithms, training data quality, or latent test-time reasoning rather than enlarging the base network size.

\secpar{Alibaba}
The Qwen closed-weight models produce bounds that strictly follow their designated tiers. The flagship \texttt{qwen3.6-max-preview} establishes a lower bound of 104B. Below it, the flash variants decrease accordingly: \texttt{qwen3.6-flash} yields 35B, and the earlier \texttt{qwen3.5-flash-02-23} sits at 27B. Similar to OpenAI, the Qwen flagship operates with an estimated minimum size kept near 100B, maintaining a relatively small operational footprint.

\vspace{10pt}
\noindent
Across all evaluated providers, the extracted size signals behave systematically. Within any single generation, the inferred bounds correctly respect the provider's product hierarchy (e.g., Opus $>$ Sonnet $>$ Haiku) without this ordering being provided to the methodology. Furthermore, the data reveals a clear divide in the industry's approach to scaling. Google and Anthropic exhibit behavior consistent with large increases in parameter counts or dense-equivalent routing, establishing minimum sizes exceeding 400B for their primary models. Conversely, OpenAI and Qwen operate with flagships bounded near 110B, suggesting a strategy that halts parameter growth to focus on optimizing a fixed model capacity. Assuming models from different developers memorize text similarly, these divergent bounds highlight that advanced capabilities are currently being achieved through two distinct architectural philosophies.

\section{Related Work}
\label{sec:related_work}

A substantial body of work has studied memorization as an intrinsic property of large language models. Carlini et al.~\cite{carlini2021extracting} demonstrated that training data can be directly extracted from language models through targeted prompting, establishing that verbatim memorization of text is a measurable and systematic phenomenon rather than an incidental artifact. Subsequent work by Carlini et al.~\cite{carlini2022quantifying} quantified how memorization scales with model size, data repetition, and context length, showing that larger models memorize training data more extensively—a relationship our method exploits as a size signal. Tirumala et al.~\cite{tirumala2022memorization} further characterized the dynamics of this effect, finding that memorization emerges early in training and increases monotonically with parameter count, without necessarily causing overfitting. Collectively, this work provides the empirical and theoretical grounding for our core assumption that next-token accuracy on popular, widely-circulated texts is dominated by model size.

Historically, the exact parameter counts of closed-weight models have been estimated through inference economics rather than behavioral probing \cite{erdil2024frontier, cai2504you}. These approaches attempt to reverse-engineer model size using external metadata, such as API pricing, token throughput, and latency variations, or through behavioral probes of API outputs. In practice, however, neither yields reliable estimates: economics-based approaches depend heavily on assumptions about provider hardware and pricing strategies, while output-based probes require full log-probability access that most commercial providers do not expose.

Recently, Li \cite{Li2026IKP} directly addressed the problem of LLM size inference. Grounded in the insight that factual knowledge capacity cannot be compressed and is bounded by parameter count, they introduce ``Incompressible Knowledge Probes'' (IKPs). While both their work and ours share the underlying premise that memorization provides a mathematical bound on model size, the two papers diverge significantly in signal extraction, statistical methodology, and empirical claims.

To measure capacity, Li extracts a size signal through factual question-answering. They construct a tiered benchmark of 1,400 questions sourced from LLM generation, Wikidata, and academic databases, which requires  manual auditing to resolve name collisions and grounding ambiguities. The target models' answers are then graded by an external LLM judge using a multi-way rubric (e.g., strong correct, weak correct, refusal, wrong). Wrong answers are assigned a specific numerical penalty empirically tuned to maximize the $R^2$ fit on a calibration set. This approach has notable strengths: because it relies on semantic LLM grading rather than strict output formatting, Li can evaluate models even when reasoning or chain-of-thought generation cannot be disabled, allowing them to test a large number of models. Furthermore, their log-linear regressions yield exact point estimates for effective parameter capacity. 

In contrast, our framework relies on continuous, verbatim next-token prediction measuring the recall of widely circulated, public-domain texts. This isolates intrinsic memorization without the layer of semantic interpretation, eliminating the need for an LLM judge, heuristic scoring rubric, or additional probe-filtering pipelines. By operating directly on raw, unedited canonical texts, our signal extraction avoids the risk of overfitting to specific benchmark formats or carefully tuned penalty parameters. Rather than immediately collapsing performance into a single scalar score prior to analysis, we aggregate behavioral measurements across varying prefix lengths into a high-dimensional accuracy profile vector. On top of this, we employ well-established techniques---specifically, dimensionality reduction via Principal Component Analysis (PCA) to extract a latent index $z_M$, alongside rigorous blocked sign-permutation tests---to isolate the size-dependent signal.

Consequently, the expected outputs of the two frameworks differ in scope and strength. Li's methodology provides both absolute point estimates and prediction intervals, explicitly noting that their figures for heavily safety-tuned models function as lower bounds. Their setup is more heuristic but highly versatile, allowing it to trace capacity across a much larger volume of models to provide comprehensive size profiles. In contrast, our approach is intentionally more conservative; we forgo point estimation entirely in favor of deriving more formal lower bounds through more rigorous aggregation methods.

Empirically, these differences in signal extraction and methodology lead to diverging outcomes. 
Overall, the lower bounds we obtain for closed-weight models are far more conservative than the figures reported by Li. 
As a concrete illustration, our lower bound for OpenAI GPT Mini is 12B parameters, while Li's framework reports 137B for the same model—a gap of an order of magnitude. 
Furthermore, the two approaches can yield conflicting generational trajectories. For instance, between Anthropic's Claude 3.5 Haiku and Haiku 4.5, Li's framework registers a decrease (more than halved) in apparent size---a known artifact where stricter safety and refusal policies are amplified by their explicit scoring penalty. Conversely, our framework produces a larger lower bound between these exact same generations.

\section{Conclusions}

We introduced a novel, black-box methodology to infer conservative lower bounds on the size of LLMs. The primary goal was to isolate an objective, stable signal of model size that resists overfitting. As the signal, we used the next-word prediction accuracy on memorized popular texts. Such aggregated signals over different texts and prefix lengths resulted in high-dimensional accuracy profile vectors. Our dual-inference framework---combining a PCA driven scaling-law estimator, and a statistical hypothesis test---translates this memorization signal into conservative lower bounds.
Evaluated against a broad spectrum of open-weight dense and Mixture-of-Experts (MoE) models, our method  yielded valid lower bounds. When extended to closed-weight systems, the inferred bounds  recovered internal product hierarchies and generational scaling shifts within developers like Anthropic and Google. This demonstrates that a closed model's  size can be systematically probed even under extremely limited, black-box API conditions.

Despite these promising results, size inference remains subject to significant challenges. The bounds produced are conservative by construction, and closing the gap between the inferred lower bound and the true model size remains an open problem. Additionally, MoE architectures further obscure size inference, as only a fraction of parameters is engaged on any given input, making it difficult to recover the true total parameter count from observed outputs alone. Addressing both limitations is the primary direction for future work.

\bibliographystyle{IEEEtran}
\bibliography{paper}


\appendices

\section{Texts and Prompts}
\label{sec:appendix:texts-and-prompts}
\begin{table}[H]
\caption{Popular texts used in the evaluation.}
\label{tbl:novels}
\centering
\scriptsize
\setlength{\tabcolsep}{3pt}
\begin{tabular}{p{0.9\columnwidth}}
\toprule
Title \\
\midrule
Alice in Wonderland \\
Frankenstein \\
Grimms Fairy Tales \\
Hamlet \\
Julius Caesar \\
Macbeth \\
Moby Dick \\
Moonstone \\
Oliver Twist \\
On the Origin of Species \\
Peter Pan \\
Pride and Prejudice \\
Psalms \\
Robinson Crusoe \\
Romeo and Juliet \\
The Adventures of Sherlock Holmes \\
The Analects of Confucius \\
The Art of War \\
The Bhagavad Gita \\
The Bible \\
The Bill of Rights \\
The Book of Mormon \\
The Charter of the United Nations \\
The Cloister and the Hearth \\
The Communist Manifesto \\
The Declaration of Independence \\
The Gospel According to Saint Matthew \\
The Great Gatsby \\
The Magna Carta \\
The Odyssey \\
The Picture of Dorian Gray \\
The Prince \\
The Republic \\
The Sermon on the Mount \\
The United States Constitution \\
The Universal Declaration of Human Rights \\
The Wonderful Wizard of Oz \\
\bottomrule
\end{tabular}
\end{table}

\begin{table}[!h]
\caption{Prompt variants (consisting of system and user prompts) used for next-word prediction. The phrase ``be or not to be that is the'' is shown here only as an illustrative example and is not necessarily used in the actual tests. It could have appeared as one of the incomplete sentences from Hamlet, corresponding to prefix length of 8 (as it has 8 words).
}
\label{tbl:prompts}
\centering
\scriptsize
\setlength{\tabcolsep}{3pt}
\begin{tabular}{p{0.95\columnwidth}}
\toprule
\textbf{Prompt 1}\newline \textbf{System:}\newline {\ttfamily You are next word prediction model, given a phrase you output the next word.\newline You always output a single word without spaces or punctuation.\newline Only respond with the next word, without any additional text or explanation.\newline If you are unsure, provide your best guess.}\newline \textbf{User:}\newline {\ttfamily "be or not to be that is the"} \\
\midrule
\textbf{Prompt 2}\newline \textbf{System:}\newline {\ttfamily \textit{None}}\newline \textbf{User:}\newline {\ttfamily Output only one next continuation word for this phrase "be or not to be that is the"} \\
\midrule
\textbf{Prompt 3}\newline \textbf{System:}\newline {\ttfamily \textit{None}}\newline \textbf{User:}\newline {\ttfamily Text: "be or not to be that is the"\newline What is the precise next word that comes immediately after the text above? Reply with strictly one word.} \\
\midrule
\textbf{Prompt 4}\newline \textbf{System:}\newline {\ttfamily \textit{None}}\newline \textbf{User:}\newline {\ttfamily Context: "be or not to be that is the"\newline Instruction: Provide the immediate next word. Limit your output strictly to exactly one word.} \\
\midrule
\textbf{Prompt 5}\newline \textbf{System:}\newline {\ttfamily \textit{None}}\newline \textbf{User:}\newline {\ttfamily [BEGIN PREFIX]\newline be or not to be that is the\newline [END PREFIX]\newline Return the next word only.} \\
\bottomrule
\end{tabular}
\end{table}

\section{Correctness of Assumption~1}
\label{sec:appendix:assumption-1}

Our Assumption~1 states that sufficiently popular texts appear with approximately uniform frequency across all pretraining corpora. Since the underlying training data of these models is inaccessible, we cannot confirm this claim directly. Instead, we test the closest observable alternative: whether all models agree on which texts they predict more accurately than others. If Assumption~1 holds, per-text accuracy should be a stable property of the text itself, independent of which model is performing the prediction.

For each model, we compute its mean accuracy per text, averaged across all tested prefix lengths. We then convert these values to ranks, assigning $1$ to the text with the highest mean accuracy and $|\mathcal{T}|$ to the text with the lowest. To measure the agreement between any two models' ranked lists, we utilize the Spearman rank correlation $\rho \in [-1,\, 1]$. A value of $\rho = 1$ indicates that both models rank the texts in exactly the same order, $\rho = 0$ indicates zero correlation, and $\rho = -1$ indicates perfectly reversed orderings. We intentionally operate on ranks rather than raw accuracy values; because a larger model is uniformly more accurate on all texts, raw-value agreement would predominantly reflect similarity in model size rather than shared text exposure.

Across all $\binom{56}{2} = 1540$ possible model pairs evaluated on the 37 popular source texts, the mean rank correlation is $\bar{\rho} = 0.906$, with $86.4\%$ of pairs exceeding $\rho = 0.80$ (a threshold generally regarded as indicating strong agreement). Even the lowest observed correlation among any pair remains notably high at $\rho_{\min} = 0.603$. This demonstrates that models consistently rank popular texts by accuracy in the exact same pattern, regardless of their size, underlying architecture, or developer. Furthermore, evaluating these groups in isolation yields similarly high intra-group consistencies: open-weight dense models achieve a mean correlation of $\rho = 0.883$, open-weight MoE models reach $\rho = 0.955$, and closed-weight models attain $\rho = 0.904$.

To verify that this overwhelming agreement is specifically driven by uniform training exposure---and is not merely an artifact of our evaluation framework---we repeat the identical test on the four obscure \emph{baseline texts}. Here, the mean correlation drops precipitously to $\bar{\rho} = 0.428$, with pairwise measurements falling as low as $\rho = -0.949$. This high variance and lack of consensus is precisely what is expected when models have encountered documents in minimal and essentially random amounts. We statistically confirm the separation between the popular and baseline text correlation distributions using a Mann-Whitney $U$ test. This distribution-free test which evaluates whether values from one group are systematically larger than the other yields $p = 3.0 \times 10^{-99}$, effectively ruling out chance and strongly validating our assumption of uniform exposure for popular texts.

\end{document}